\def\year{2022}\relax
\documentclass[letterpaper]{article} 
\usepackage{aaai22}  
\usepackage{times}  
\usepackage{helvet}  
\usepackage{courier}  
\usepackage[hyphens]{url}  
\usepackage{graphicx} 
\usepackage{natbib}  
\usepackage{caption} 
\DeclareCaptionStyle{ruled}{labelfont=normalfont,labelsep=colon,strut=off} 
\usepackage{algorithm}
\usepackage{algorithmic}

\usepackage{newfloat}
\usepackage{listings}
\floatstyle{ruled}
\newfloat{listing}{tb}{lst}{}
\floatname{listing}{Listing}
\title{Improving Sample Efficiency of Value Based Models Using Attention and Vision Transformers}
\author{
    Amir Ardalan Kalantari\textsuperscript{\rm 1}\textsuperscript{\rm 2}\thanks{Corresponding Author @  amir.kalantaridehaghi@mail.mcgill.ca}, 
    Mohammad Amini\textsuperscript{\rm 1}\textsuperscript{\rm 2}, 
    Sarath Chandar\textsuperscript{\rm 2}\textsuperscript{\rm 3}, 
    Doina Precup\textsuperscript{\rm1}\textsuperscript{\rm 2}
}
\affiliations{
    \textsuperscript{\rm 1} McGill University\\
    \textsuperscript{\rm 2} Mila \\
    \textsuperscript{\rm 3} Ècole Polytechnique de Montrèal 

}

\usepackage{bibentry}

\begin{document}

\maketitle

\begin{abstract}
Much of recent Deep Reinforcement Learning success is owed to the neural architecture's potential to learn and use effective internal representations of the world. While many current algorithms  access a simulator to train  with a large amount of data,  in realistic settings, including while playing games that may be played against people, collecting experience can  be quite costly. In this paper, we introduce a deep reinforcement learning architecture whose purpose is to  increase sample efficiency without sacrificing performance. We design this architecture by incorporating advances achieved in recent years in the field of Natural Language Processing and Computer Vision. Specifically, we propose a visually attentive model that uses transformers to learn a self-attention mechanism on the feature maps of the state representation, while simultaneously optimizing return. We demonstrate empirically that this architecture improves sample complexity for several Atari environments, while also achieving better performance in some of the games. 
\end{abstract}

\section{Introduction}

Representation learning is a critical ingredient of successful deep reinforcement learning (RL) algorithms~\citep{stooke2021decoupling}. Our goal in this paper is to investigate neural architectures that can reduce the amount of data necessary to train deep RL agents, without sacrificing performance.
Our approach  is to map the high dimensional observation space to low dimensional latent feature maps and then develop a Vision Transformer based self-attention mechanism that works on top of the latent feature maps.
We evaluate our method by running extensive experiments on 18 games from the Atari 2600 Arcade Learning Environment~\citep{bellemare2013arcade}. Our approach improves  sample efficiency in most of these games, while also achieving  higher expected return in several games as well. For this study, we consider Deep Q-Networks (DQNs) \citep{mnih2015human} which combine Q Learning with convolutional neural networks and experience replay buffer to allow an agent to learn from raw pixel observations. After 200 millions frames of experience, the DQN agent could perform on a human level on many of the Atari 2600 games. Our goal is to improve the sample efficiency of DQN by using vision transformers.

\section{Transformers}




Transformers were introduced first by \cite{vaswani2017attention}. 
The Transformer architecture uses a multi-head scaled dot-product attention mechanism, originally developed for use in natural language processing (NLP), specifically for the task of machine translation. The  attention function receives three inputs of key ($K$), value ($V$) and query ($Q$).
The queries and keys have dimension  $d_k$ and the values have dimension $d_v$.
The attention is calculated as:
\begin{equation}
    Attention(K,V,Q) = softmax(\frac{QK^T}{\sqrt{d_k}})V
\end{equation}
The architecture is said to be multi-head because it linearly projects multiple sets of key, query and value in parallel and concatenates them together at the end. The authors argue that this architecture enables the model to use different subspaces to maximize its objective.

\begin{figure}[h!]
\centering
\includegraphics[page=1,width=0.4\textwidth]{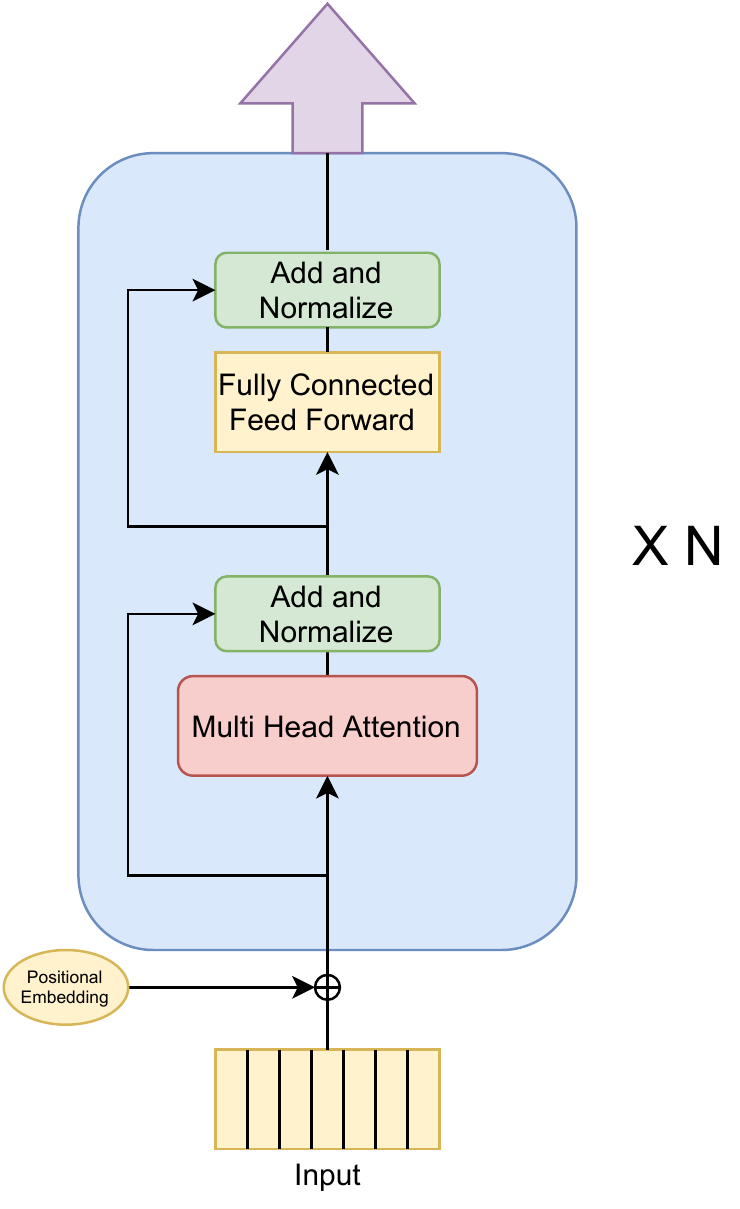}
\caption{Transformer Encoder.}
\label{fig:Encoder}
\end{figure}

The Transformer architecture ended up revolutionizing many sequence learning tasks in NLP and beyond,  rendering  previous models obsolete and removing the need for recurrence altogether in many cases.  The Encoder of the Transformer architecture  used in our model is depicted in Figure-\ref{fig:Encoder}.

\begin{figure}[h!]
\centering
\includegraphics[page=1,width=0.5\textwidth]{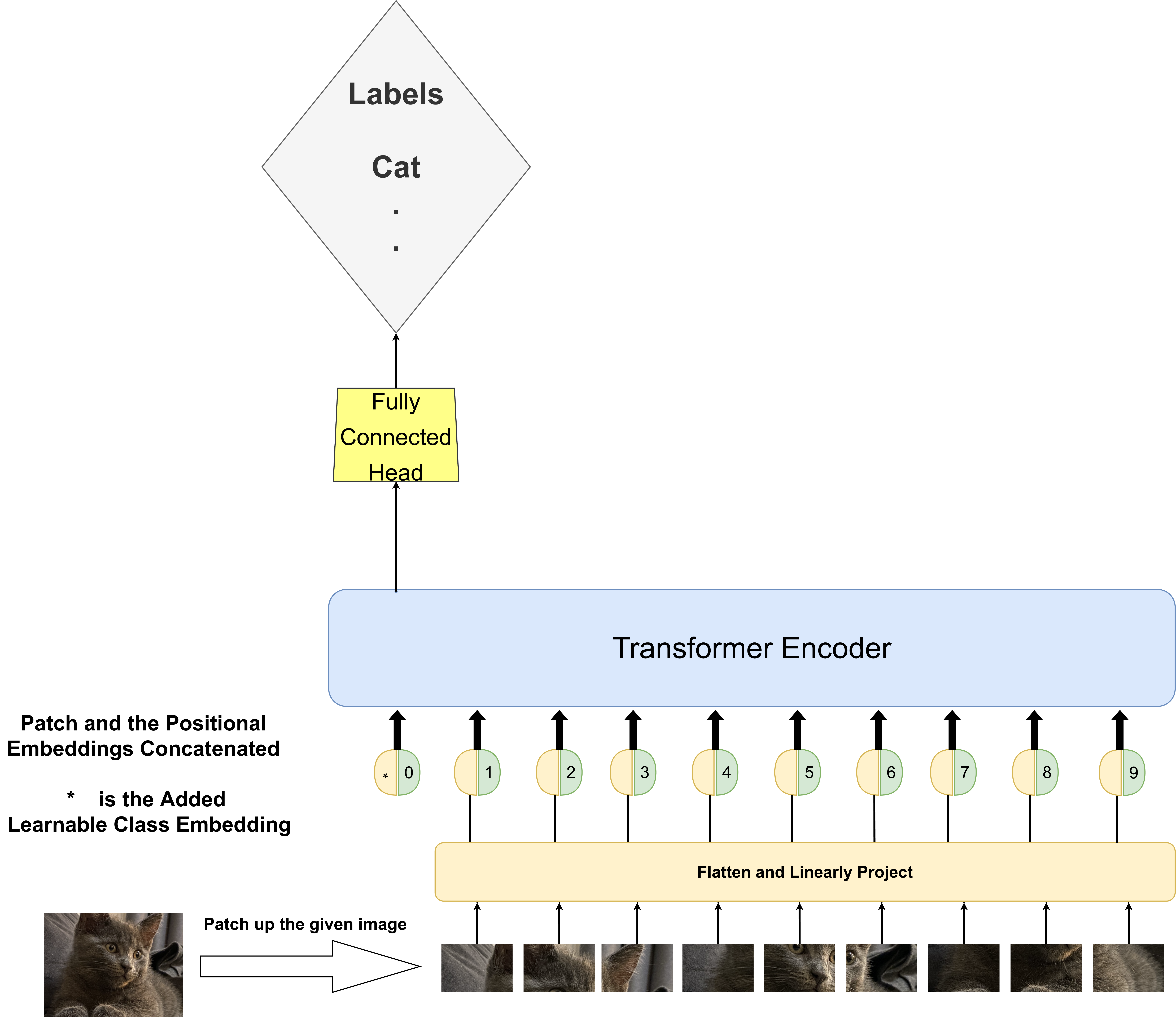}
\caption{Vision Transformer Architecture}
\label{fig:ViT}
\end{figure}

More recently,  we have been witnessing a shift from CNNs to Transformer based models in the field of computer vision as well. Transformers have been used in the task of image generation on datasets like ImageNet \citep{parmar2018image}, but the first  major success  was the Vision Transformer (ViT) proposed by \cite{dosovitskiy2020image}. Figure\ref{fig:ViT} depicts the structure of this type of model. In  ViT, the model does not use any convolutional filters. Instead, the image is first split into patches, and the patches are flattened. Afterward, the patches go through a fully connected linear layer. Once the patches are projected to get the Pixel Embeddings, they are concatenated with a \textit{Positional Embedding}. The necessity of a Positional Embedding in Transformers arises from the fact that the model uses no recurrence or convolutional layers. Hence, the model has no way to deduce the order of tokens and inputs, since they are all fed into the network simultaneously. In the case of ViT, without the positional embeddings, the model wouldn't know the relative or absolute order of the patches with respect to the original image. The choice of  Positional Embeddings is an active area of research by itself \citep{luo2021stable,ke2020rethinking}. 
The ViT also adds an extra learnable class embedding, which is used to extract the class labels at the end of the network.

Once the pixel embeddings are concatenated with their corresponding positional embeddings, they are fed into a Transformer encoder. The encoder has a fully connected layer at the end, from which the classes are extracted. When trained solely on medium-sized databases such as ImageNet, ViT performs sub-optimally. The authors argue that this may be due to the lack of inductive biases which are present within the CNN architecture. However, when they train the ViT model with significantly large databases, they start to gain a significant advantage over  CNN-based models. ViT uses a Fully Connected MLP head to extract the class labels during  pre-training, and switches to a single linear layer during fine-tuning.


\section{Related Work}
Reinforcement learning has had great success after combining powerful functional approximators such as neural networks with RL algorithms, however, there is still plenty of room for improvement. 
Recently,  representation and training methods that can produce better sample efficiency for RL have been gaining  attention. We mention below the works most relevant to this paper.

\citet{laskin2020reinforcement} proposed data augmentation to improve data-efficiency and generalization.
\citet{sorokin2015deep} proposed a  deep recurrent neural network that uses attention mechanism to train a Q-learning agent. \citet{mnih2014recurrent} used a recurrent neural network coupled with a CNN for visual attention on tasks of image recognition.
\citet{gregor2018visual} used \textit{glimpse}, a \textit{k}-square body of patches which is used by an agent to attend while playing Pac-Man.
 \citet{yuezhang2018initial} used an attention mask on the features extracted by the convolutional neural network backbone for the purpose of feature selection.
\citet{chen2021decision}  proposed the Decision 
Transformer, an architecture that allows  treating the problem of learning to control an environment as a sequence modeling problem, in which the output is the action to be taken at each time step. They use collected experiences to train the model in an offline fashion.  
\citet{zambaldi2018deep} used Relational Module, a Multi Head Attention followed by feature-wise max pooling layer, to adopt relational inductive biases for Box-World and StarCraft II games. Work by \citet{parisotto2020stabilizing} proposes a  transformer based architecture to achieve more stability in RL settings. Another work by \citet{parisotto2021efficient} proposed a novel distillation method from actor to learner in the distributed RL training, this work utilized transformers as a form of model compression between separate actor and learner models.

\begin{figure*}
\centering
\includegraphics[width=0.85\textwidth]{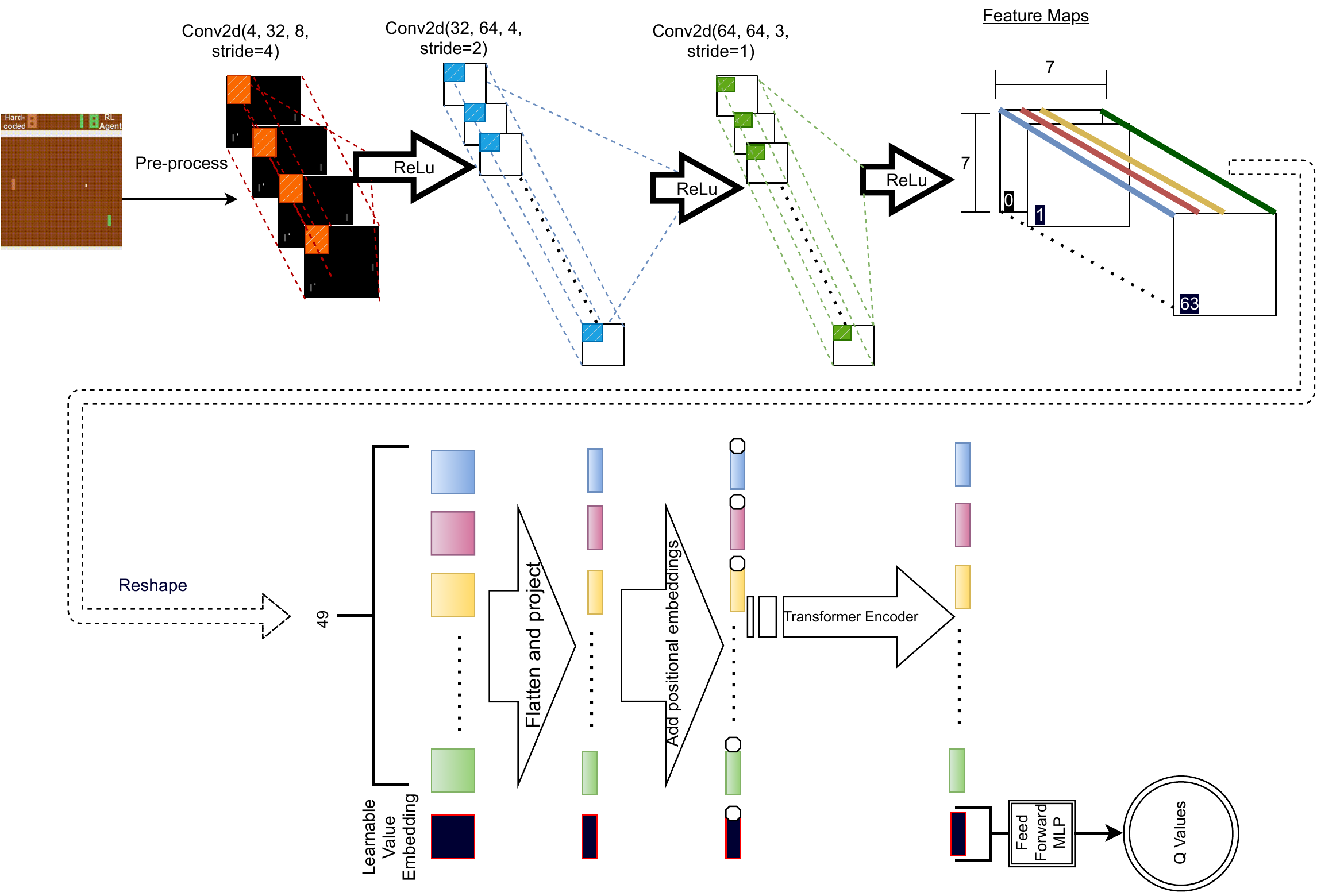}
\caption{Illustration of the architecture used.}
\label{fig:architecture}
\end{figure*}
\section{Methodology}

\begin{figure*}
\centering
\includegraphics[width=1.0\textwidth]{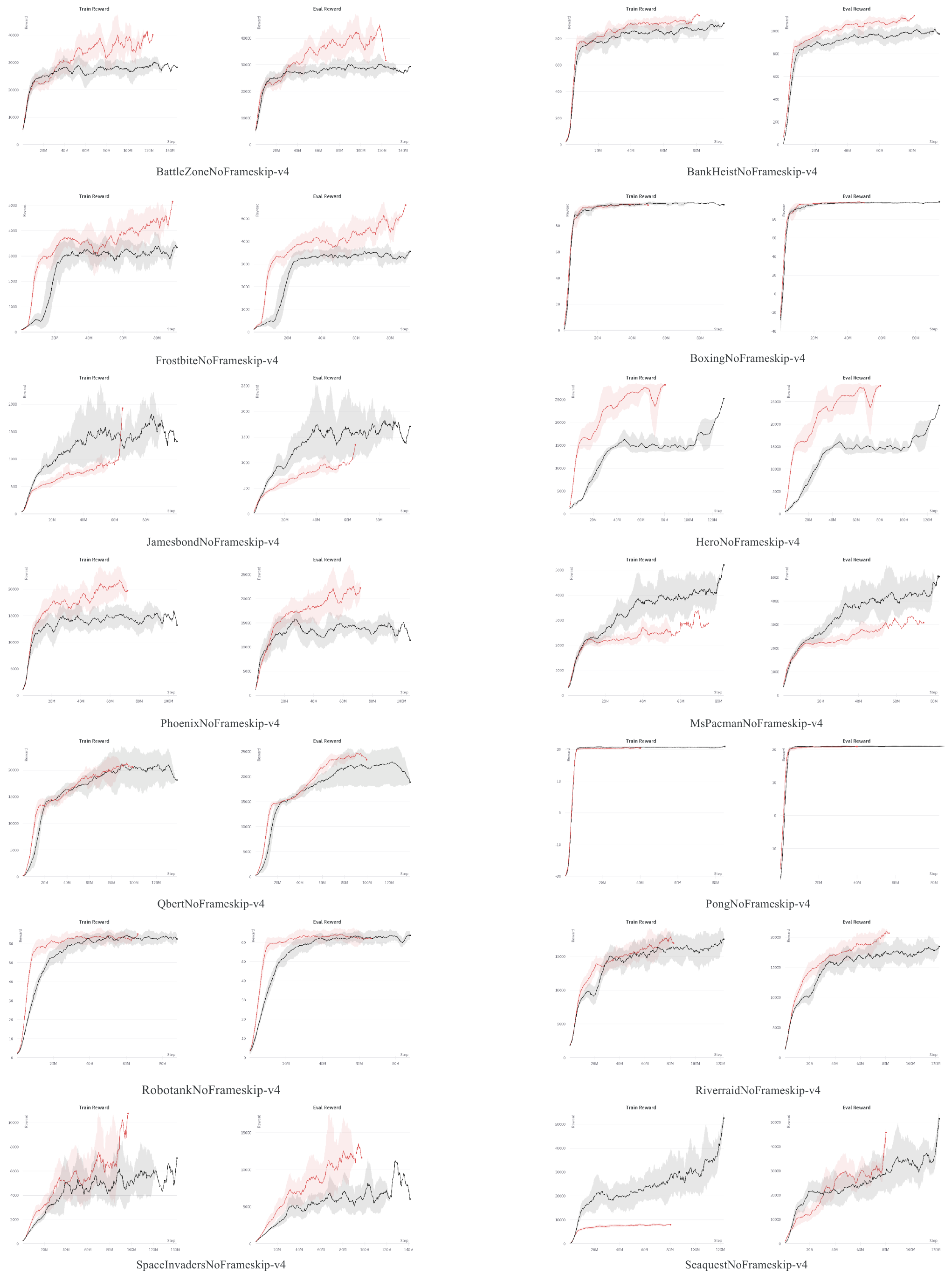}
\end{figure*}

\begin{figure*}
\centering
\includegraphics[width=1.0\textwidth]{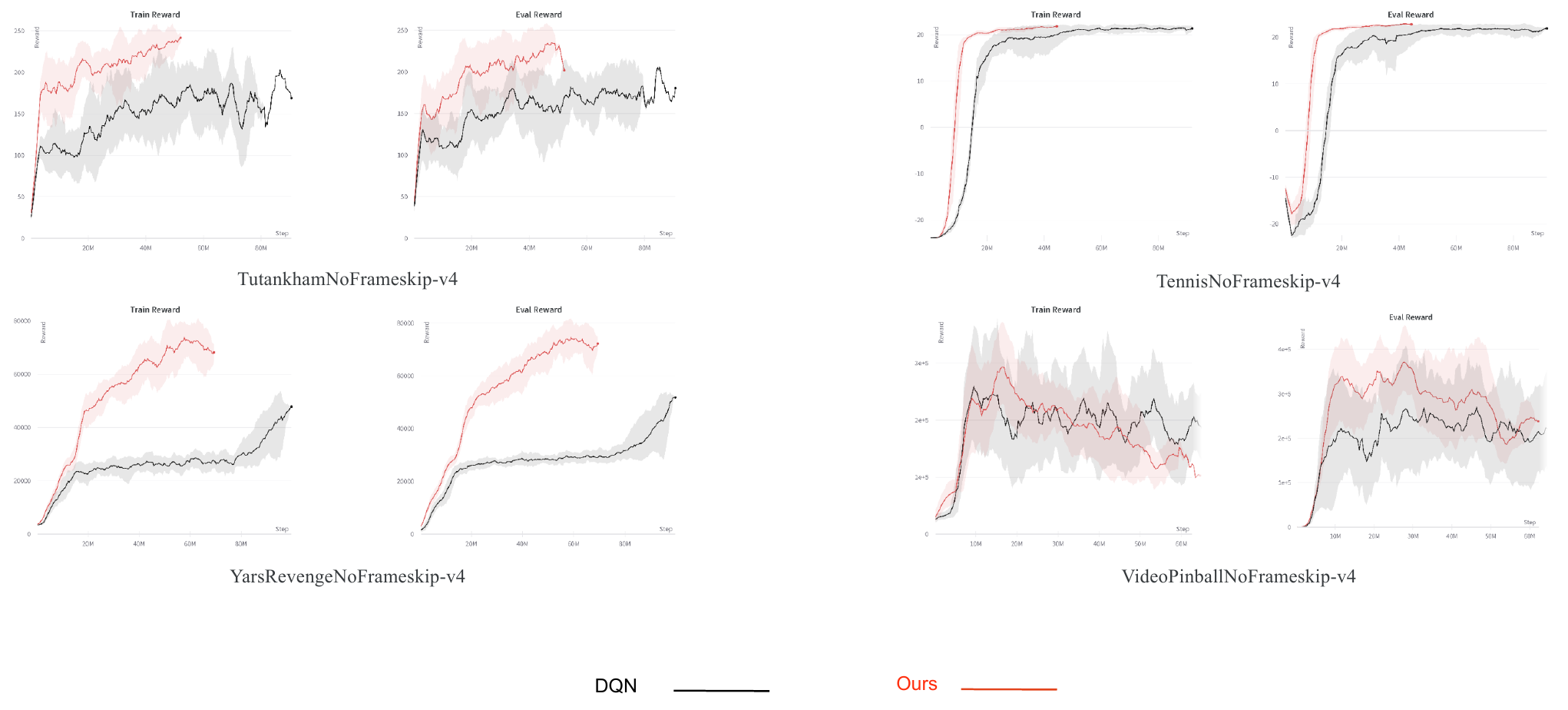}
\caption{Training and Evaluation graphs for our method. }
\label{fig:published_individual}
\end{figure*}

Our methodology is based on the model developed in the original Deep Q-learning paper~\citep{mnih2013playing} and recent developments in Transformer based attention models. 
We described it here in the context of the Atari environment in which we carry out our experiments. In this setup, image frames are pre-processed by being resized to 84x84 pixels and grayscaled. Four consecutive frames are stacked to reduce  partial observability, as a single frame cannot convey the direction, speed, and acceleration of the moving elements on the screen. The stacked-up image frames are then passed through a CNN (with filters and dimensions stated in figure \ref{fig:architecture}), through which features are extracted from the input, and subsequently the feature maps are flattened and passed through a fully-connected layer, and afterward the action values are provided at the output. Q-learning is used to train the network, using a replay buffer of sampled transitions, in order to ensure that the data appears more i.i.d to the network.  

Our proposed method follows the same architecture until the feature extraction step as depicted in Figure \ref{fig:architecture}. 
We use a CNN with the same exact architecture and dimension to extract 64 feature maps from the input. Each feature map has dimension \textbf{7} by \textbf{7}. We then proceed to reshape the feature maps to get 49 features of dimension 64, as follows:
\begin{equation}
    \textrm{Channel} * \textrm{Height} * \textrm{Width}   \rightarrow \textrm{(Height * Width)}  * \textrm{Channel}
\end{equation}

We flatten and linearly project the new features using a fully connected layer to get a feature embedding. We then add a learnable positional embedding to each one of the features, similarly to the ViT architecture. We also add a \textit{learnable value embedding}, along with its corresponding positional embedding. The purpose of this additional embedding is to extract the relevant information regarding the Q-values after the final layer of the Transformer encoder. 
We use a latent dimension of 128 to project our feature tokens onto. The \textit{learnable value embedding} also has the same dimension as the rest of the input tokens.  The positional embeddings likewise have a latent dimension of 128 and they are concatenated with their corresponding feature embeddings. 

Following this, the inputs are passed through a Transformer encoder. We used the Linformer architecture \citep{wang2020linformer} to save compute time and resources during our experiments. At the end of the encoder, we discard all the tokens except the value token, which we pass through a LayerNorm followed by a feed-forward MLP head of size 128 $\rightarrow$ \textit{Num actions}, \textit{Num actions} being the number of actions the agent can take in the environment, to arrive at the Q-values.
The reported results use an 8 head, 2 layered transformer encoder on top of the CNN. For exploration, we used an annealing $\epsilon$-greedy algorithm starting from a value of $1.0$ and linearly decreasing to $0.01$. 

We benchmarked our method on 18 of the ALE Atari games against a baseline  DQN algorithm~\citep{mnih2015human}. We used dropout regularization~\citep{srivastava2014dropout} with a dropout rate of 0.1.
For the baseline DQN, we used the previous reported hyper-parameters and pre-processing as to not cause any unfair advantage for our method. The list of hyper-parameters used in the baseline DQN model is listed in table 1.
\begin{table}
\begin{tabular}{lll} \hline
Hyper-parameter & Value  \\ \hline
Buffer Size & 1000000  \\ \hline
Gamma & 0.99  \\ \hline
Target Network Update Frequency & 30000 \\ \hline
Mini-batch  size & 32  \\ \hline
Optimizer learn rate & 1e-4 ADAM  \\ \hline
Optimizer Eps & 0.00015  \\ \hline
Gamma & 0.99  \\ \hline
Start Epsilon & 1.0  \\ \hline
Finish Epsilon & 0.01  \\ \hline
Exploration Fraction & 5 Million Steps  \\ \hline
Learning Starts & 200000  \\ \hline
Train Frequency & 4  \\ \hline
\end{tabular}
\caption{List of Hyper-parameters }
\label{table:1}
\end{table}
We used the same hyper-parameters for our model, with the exception of the batch size (which we discuss in more detail below).

\section{Results and Discussion}

Figure 4 summarizes the empirical results. Each pairs of plots corresponds to a training and evaluation run, plotting return as a function of the number of time steps.  We ran each experiment using three seeds and averaged the return over these. We also used a running average with a window size of 10 to smooth out the graphs. We used Weights \& Biases \cite{wandb} for experimental tracking and visualization.  For every 10 episodes of training, we ran 2 evaluation episodes, during which we turned off the dropout and epsilon-greedy exploration, as well as the learning, and we recorded the return of the current learned policy.  The training curves are plotted on the left column of each game, and the evaluation returns are plotted in the right column for each game.
The solid line in the graphs corresponds to the mean of the experiment and the shaded areas correspond to the minimum and the maximum return obtained.
For the sake of consistency, we ran our own baseline DQN instead of comparing to previously reported results. It is worth mentioning that our baseline DQN performs better than previously reported DQN results in most of the games.

Running the experiments on the entire Atari set using 200,000,000 steps was not  feasible for us. This is due to the fact that although our method uses fewer parameters than the baseline DQN, it takes approximately 3-4 times longer to train. Hence, we ran the experiments until the training seemed to converge, or until the relative ordering of the methods became apparent. Of the 18 games we tried, our method performed \textit{better} than the DQN baseline in 15 of them. We define ``better" as either  having converged to a higher return plateau, or converging to the same return level but with fewer samples. In some of the games,  such as \textit{HeroNoFrameskip-v4},\textit{FrostbiteNoFrameskip-v4} and \textit{YarsRevengeNoFrameskip-v4}, the gain of our method in terms of sample efficiency is quite large.

\begin{figure}
\centering
\includegraphics[width=0.5\textwidth]{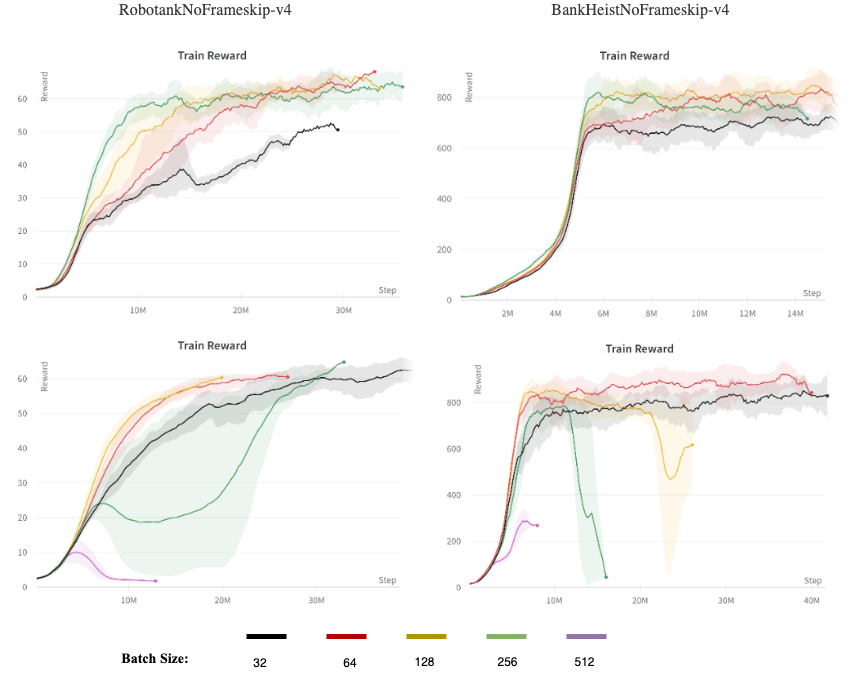}
\caption{Batch size effect}
The top row shows the effect of increasing batch size on our method in two environments. 
The bottom row corresponds to the baseline DQN method.
\label{fig:ablation}
\end{figure}

Our method underperforms in three of the 18 games, namely \textit{MsPacman}, \textit{Jamesbond} and \textit{Seaquest}. In the curious case of Seaquest, after around 40,000,000 steps, our method seems to catch up and surpass the baseline DQN on evaluation episodes. We took our saved model and let it play the game for some episodes both with and without dropout, to understand what is happening. It appears that the difference between the performance in the training episodes compared to the testing ones depends on how the agent handles the case in which its oxygen runs low. When there is no dropout, the agent seems to handle the low oxygen situations by staying just below the surface to avoid dying, while dropping in depth by only slightly in order to shoot down enemy and accumulate the 10000 points to afford another life. However, when dropout is activated, the agent acts sub-optimally in situations with low oxygen, either crashing into objects or surfacing without having any divers and dying off. We suspect that this is due to the randomness introduced by the dropout.

We initially tried an alternative end-to-end attention based model, such as the ViT. However the fully attentive models took significantly longer to train, which made them impractical for almost all the games, with a few exceptions, such as Pong. These models were harder to stabilize and get to work, and during our limited experiments, we noticed that they were less sample efficient than the model depicted in Figure \ref{fig:architecture}. This might be due to the fact that end-to-end attentive models have fewer inductive biases. This  observation is consistent with the results in the original ViT paper, since the model evaluated there required immense amounts of data to outperform the baseline methods, which benefitted from the inductive bias of CNNs. One of the most important factors in stabilizing an end-to-end attentive model was the mini-batch size at each learning stage. When using the standard mini-batch size of 32, the model failed to learn, with returns resembling a sinusoidal function, as a function of the number of time steps. By increasing the batch size to 256, the network stabilized and started to learn. We kept this setting for our model throughout our experiments. 

In graph \ref{fig:ablation}, we show the effect of varying mini-batch size on the CNN and transformer hybrid method versus the baseline DQN. In the top two plots of these experiments, we observe a consistent gain in the performance of the visually attentive model as the mini-batch size increases. The bottom two plots show the effect of various mini-batch sizes on the performance of the baseline DQN. In the case of the baseline DQN, increasing the mini-batch size to a small extent seems to improve the performance marginally, however, increasing it to 128 and beyond seems to introduce instability and leads to a degradation in the quality of the model. This observation is consistent with \citep{keskar2017largebatch} and \citep{masters2018revisiting} which concludes that using a small mini-batch size yields more optimal results and are better suited for generalization. However, it seems like visually attentive models have a higher tolerance with regards to this phenomenon and could make use of computational parallelism. On this note one could conclude that while currently using visually attentive models and larger mini-batch sizes could come at a high compute cost, as compute power improves, larger mini-batch size is needed to scale faster converging DRL agents.

CNNs have been designed to exploit spatial invariance as an inductive bias. In recent works \citep{elsayed2020revisiting}, it has been shown that easing the spatial invariance bias can actually help improve classification accuracy on small CNNs. CNNs can also suffer from difficulty in capturing long-range dependencies within the input frames, in the context of RL. The proposed architecture can alleviate this  to certain extent, by having the attention mechanism applied in  feature space and capturing dependencies  between different entities in this  space. \citet{raghu2021vision} showed that in the absence of copious amounts of training data for image classification in ViTs, the models fail to develop strong local representations and receptive fields, and the attention layers only develop a global attention span.
The proposed model seems  able to get the best of both worlds in this respect. However it does come with a toll of compute time, which is 3-4 times larger than for the baseline DQN.. Hence, our proposed model would be most useful  in  situations where samples are more expensive than compute  resources. 
. 

\section{Conclusion and Future Work}
In this paper, we explored the potential of recent advancements in Transformers and Vision Transformers for building reinforcement learning agents. Based on our experiments, these types of models show great potential for improving the sample efficiency of existing reinforcement learning algorithms. While our proposed method could come at the expense of compute time with current compute power, the ability of our model to use larger batch size (i.e. 8X) assists in the scalability of DRL agents.For future work, we plan to carry out more rigorous ablation studies, improve the compute time and explore architecture choices further that allow for further stability and scalability of DRL agents. 

\section{Acknowledgment}
We would like to thank the Canadian Institute for Advanced Research (\textbf{CIFAR}) for funding this project. Also, we are very grateful for Compute Canada and its available resources and management for running the experiments pertaining this work.


\bibliography{aaai22}

\end{document}